\documentclass[letterpaper]{article} % DO NOT CHANGE THIS
\usepackage{aaai23}  % DO NOT CHANGE THIS
\usepackage{times}  % DO NOT CHANGE THIS
\usepackage{helvet}  % DO NOT CHANGE THIS
\usepackage{courier}  % DO NOT CHANGE THIS
\usepackage[hyphens]{url}  % DO NOT CHANGE THIS
\usepackage{graphicx} % DO NOT CHANGE THIS
\usepackage{natbib}  % DO NOT CHANGE THIS AND DO NOT ADD ANY OPTIONS TO IT
\usepackage{caption} % DO NOT CHANGE THIS AND DO NOT ADD ANY OPTIONS TO IT
\usepackage{algorithm}
\usepackage{algorithmic}

\usepackage{amsmath}
\usepackage{amssymb}
\usepackage{booktabs}
\usepackage{bm}
\usepackage{multirow}

\usepackage{newfloat}
\usepackage{listings}
\DeclareCaptionStyle{ruled}{labelfont=normalfont,labelsep=colon,strut=off} % DO NOT CHANGE THIS
\floatstyle{ruled}
\newfloat{listing}{tb}{lst}{}
\floatname{listing}{Listing}
\title{Mitigating Negative Style Transfer in Hybrid Dialogue System}
\author{
    Shimin Li\textsuperscript{\rm 1},
    Qinyuan Cheng\textsuperscript{\rm 1},
    Linyang Li\textsuperscript{\rm 1},
    Xipeng Qiu\textsuperscript{\rm 1,2} \thanks{Corresponding author},
}
\affiliations{
    \textsuperscript{\rm 1} School of Computer Science, Fudan University\\
    \textsuperscript{\rm 2} Shanghai Key Laboratory of Intelligent Information Processing, Fudan University\\
    \{smli20, linyangli19, xpqiu\}@fudan.edu.cn, chengqy21@m.fudan.edu.cn
}

\usepackage{bibentry}
\begin{document}

\maketitle

\begin{abstract}
As the functionality of dialogue systems evolves, hybrid dialogue systems that accomplish user-specific goals and participate in open-topic chitchat with users are attracting growing attention.
Existing research learns both tasks concurrently utilizing a multi-task fusion technique but ignores the negative transfer phenomenon induced by the unique textual style differences.
Therefore, contrastive learning based on the latent variable model is used to decouple the various textual genres in the latent space. We devise supervised and self-supervised positive and negative sample constructions for diverse datasets.
In addition, to capitalize on the style information contained in the decoupled latent variables, we employ a style prefix that incorporates latent variables further to control the generation of responses with varying styles.
We performed extensive experiments on three dialogue datasets, including a hybrid dialogue dataset and two task-oriented dialogue datasets. The experimental results demonstrate that our method can mitigate the negative style transfer issue and achieves state-of-the-art performance on multiple dialogue datasets.

\end{abstract}

\section{Introduction}
\begin{figure}[t]
\centering
\includegraphics[width=1\columnwidth]{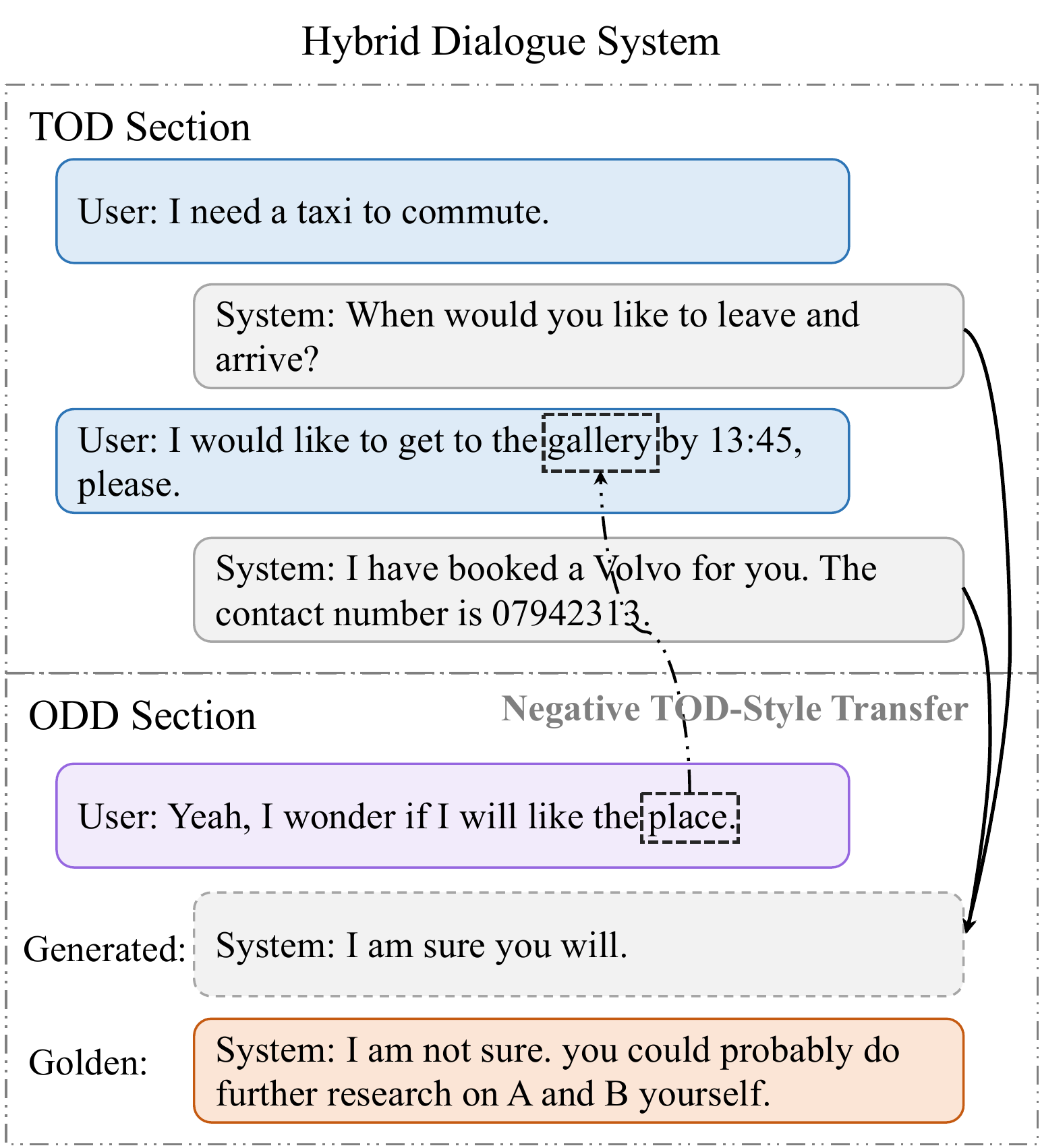} % Reduce the figure size so that it is slightly narrower than the column. Don't use precise values for figure width.This setup will avoid overfull boxes.
\caption{The different styles of responses in the hybrid dialogue system, where TOD responses are in grey, and ODD responses are in orange. The TOD style responses in the ODD section are affected, resulting in mediocre ODD responses.}
\label{task_intro}
\end{figure}

Previous research on dialogue systems has been distinctly divided into task-oriented dialogue systems (TOD) \cite{mttod, pptod, galaxy, simpletod}  and open-domain dialogue systems (ODD) \citep{blender} based on their application. Whereas task-oriented is intended for the successful completion of specific goals and instructions from the user, open-domain dialogue systems engage in open-ended chitchat with the user on various topics.
Furthermore, with the development of dialogue systems and the excellent transferability and generalization of pre-trained models to downstream tasks \citep{survey_pretrain}, there is a trend towards a progressive integration between different functional dialogue systems \citep{unids}. Some datasets that fuse various dialogue tasks have also emerged \citep{fusedchat, salesbot, accentor}.
Nevertheless, different dialogue systems vary in the textual style of their responses to facilitate the completion of a particular dialogue task \citep{stylefusion}. As illustrated in Figure \ref{task_intro}, task-oriented dialogue systems use refined discourse to complete the user's task accurately and adequately. In contrast, open-domain dialogue systems generate more varied responses to enhance user engagement and interest.

Existing hybrid dialogue systems learn different dialogue tasks through a multi-task approach or by constructing uniform data patterns \citep{pptod, unids}. However, they ignore the inconsistency in text style between task-oriented and open-domain dialogue responses, leading to the negative transfer of different dialogue tasks in the hybrid dialogue system. Even while the encoder can recognize both dialogue styles exceptionally well, the decoder cannot construct a particular style of discourse during the generation phase \cite{fusedchat}. This phenomenon is evidenced by decreased task success rates for task-oriented dialogue and vacuous generic responses for open-domain dialogue. Moreover, some datasets \citep{multiwoz} do not provide explicit text style labels or relevant indication information, rendering the modeling of text style more intractable \citep{polarized_vae}.

Motivated by the above, our proposed Hybrid-Style-Controlled Dialogue System (HiS-Dialog)\footnote{Code: \url{https://github.com/whatissimondoing/HiS-Dialog}} is designed with contrastive latent variables to model and distinguish implicit text style information. Also, the style prefix is employed to guide the dialogue system in generating responses of different styles with the modeled style information.

Firstly, to generate diverse text and model text styles, we take variational encoder-decoder (VED) \citep{vhred, dialogved} as the base architecture and a pre-trained T5 \citep{t5} as the backbone model, which maps deterministic encoder hidden states to latent variables with randomness. Then, during the response generation process, the decoder relies on the different latent variables sampled to generate various styles of dialogue responses.
However, VED alone can only model undirected and random implicit text styles, failing to steer the model to generate different response styles with directionality depending on the dialogue task \citep{polarized_vae}. 

Therefore, we propose extending the application of contrastive learning \citep{triplet_loss, simclr} to VED to enable decoupling the stylistic latent space into partially distinct textual style subspaces.
Precisely, we control the relative distances between different latent variables in the latent style space based on similarity, i.e., latent variables with similar text styles are closer in the style subspace, while the distances between different latent variables are relatively farther apart. Nevertheless, the previous work \citep{polarized_vae} only applied to cases where style labels were available. For instances without text style descriptors but with alternative subjects or domains, we developed self-supervised techniques with greater application for constructing positive and negative samples \cite{simcse}. That is, for all samples in a batch, by reparameterizing twice from its posterior distribution, two variables in the same position of two batches are regarded as positive samples, and variables in other positions are considered negative samples of each other. The method can be extended to some dialogue datasets without explicit textual style information.

Secondly, to exploit the extracted text style variables more adequately and steer the subsequent generation of responses in varying text styles, we designed style prefix as a continuous instruction incorporating style variables for controlled text generation \citep{prefix_tuning}.
Contrary to previous work, which only incorporated latent variables in the self-attention calculation at the decoder side \citep{dialogved}, HiS-Dialog adds a set of discrete-continuous vectors to both the self-attention and cross-attention calculations at the encoder and decoder. It enables the style prefix to extract more relevant information about the text style from the encoder side and steer the decoder to generate text in the target style.

We validated our approach on a hybrid dialogue dataset and two task-oriented datasets. The experimental results indicate that HiS-Dialog achieves state-of-the-art results in all three datasets.

The contributions of this paper can be summarised as follows:

\begin{itemize}
\item We explore the negative text style transfer issue in hybrid dialogue systems for the first time and propose a conditional dialogue style generation model based on contrastive latent variables.
\item Depending on the dataset's availability of response style labels, we propose both supervised and self-supervised contrastive latent variable modeling approaches.
\item We introduce a style prefix based on style variables, which empowers style information to be extracted and fully utilized by the model to steer the generation of different response styles.
\end{itemize}

\section{Related Work}
\subsection{Pre-trained Dialogue Models}
With the advancement of generative pre-training models \citep{t5, bart} and increasingly more dialog data \citep{fusedchat, multiwoz, salesbot}, end-to-end pre-trained dialog models are gaining more attention \citep{pptod, simpletod, mttod}. Much of this work addresses the adaptation of pre-trained models to downstream dialogue tasks by further pre-training the models \citep{dialogpt, dialogved} with additional dialogue data.
For instance, PPTOD \cite{pptod} extends the idea of the unified paradigm of T5 \citep{t5} to the dialogue domain by constructing prompts for further pre-training of different dialogue tasks. Regarding task format construction, work also exists to model different subtasks in dialogue modules into end-to-end form \citep{simpletod, soloist, ubar}.
In addition, works also exist that design multiple decoders for different dialogue subtasks \citep{mttod,damd, mintl}.

However, such approaches to adding decoders can significantly increase the model parameters, thus limiting scalability.
Instead, our approach perpetuates the end-to-end setting. It enables the construction of a hybrid dialogue system by introducing a small number of style prefixes without requiring further pre-training or additional decoders.

\begin{figure*}[t]
\centering
\includegraphics[width=0.95\textwidth]{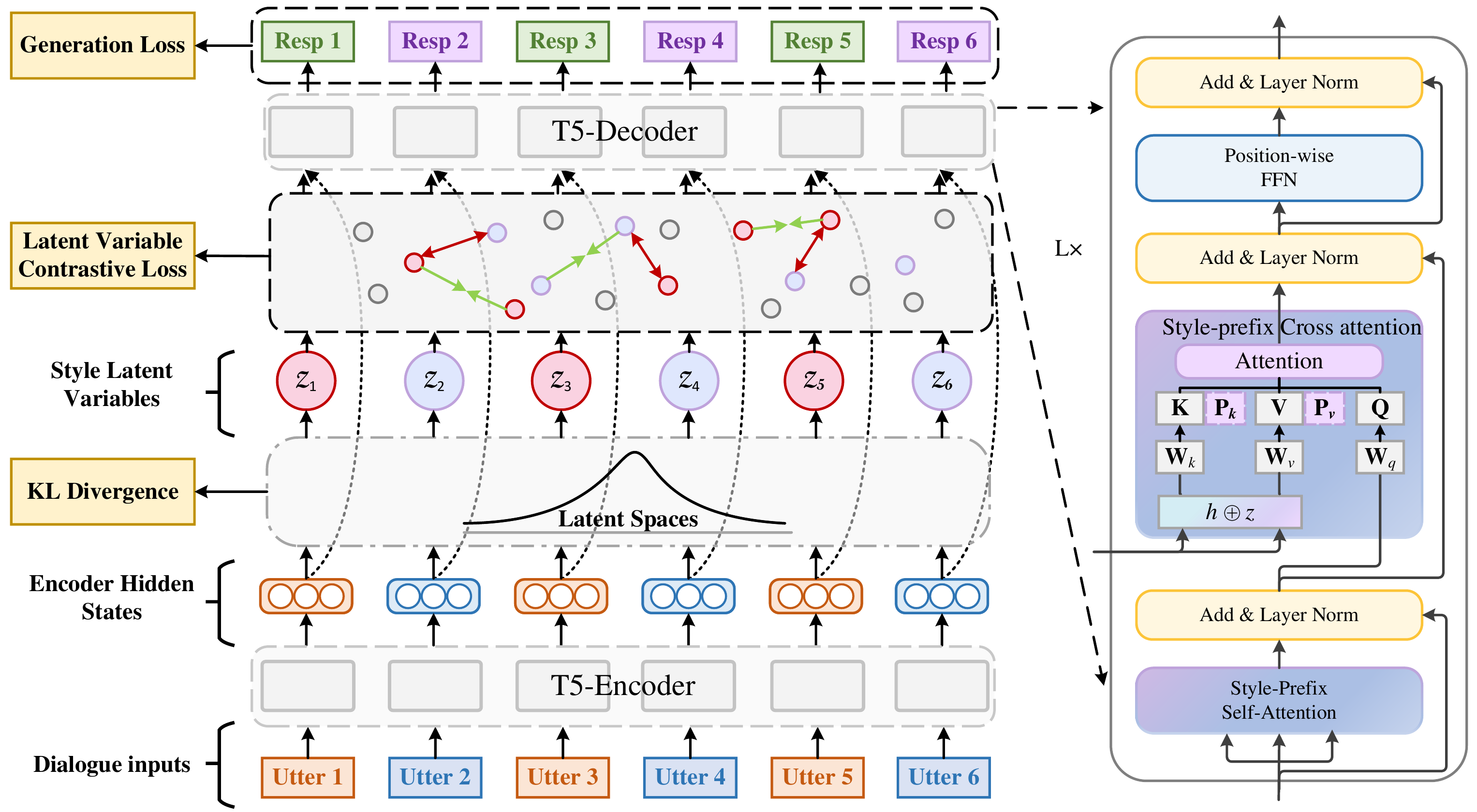} % Reduce the figure size so that it is slightly narrower than the column.
\caption{The overall framework of HiS-Dialog consists of the following steps: (1) The dialogue in a batch is encoded to obtain the hidden state and mapped to latent space for latent variables. (2) Contrastive loss is calculated to increase the distance between latent variables of different styles. (3) The style prefix combined with the latent style variable is applied to control the generation of responses with different styles.}
\label{fig:main_model}
\end{figure*}

\subsection{Hybrid Dialogue Systems}
With the gradual unification of the task paradigm \citep{para_shift}, some studies have attempted to unify several different dialogue systems. In terms of data processing, some research enables the models with some capability of chatting on top of accomplishing the task by inserting open domain dialogues into the task-oriented dialogues \citep{accentor}.
Regarding the model training approach, most existing works design multiple dialogue tasks as generation tasks and train them in a multi-task fashion \citep{pptod,unids}. However, they do not account for the relevance of contextual semantics in multiple turns of dialogue when constructing the hybrid dialogue data, making the data incompatible with real-life scenarios.
Moreover, multi-task training only extends their model's functionality but neglects the issue of negative transfer across tasks \citep{survey_transfer}. 

Unlike previous work, our model adopts a contextually coherent hybrid dialogue dataset for training while introducing contrastive loss in the latent space to mitigate the negative transfer phenomenon of multi-task training for hybrid dialogue tasks.

\section{Methodology}
This section describes the basic definition and composition of a hybrid dialogue system and how to constrain the latent space by constructing a contrastive loss on top of the latent variable encoder-decoder model. Style-controlled text generation is then conducted with style prefix incorporating the latent style variable. The overall architecture of our approach is illustrated in Figure \ref{fig:main_model}.

\subsection{Hybrid Dialogue System}
The objective is to model the task-oriented dialogue system and the open-domain dialogue system as a unified end-to-end format \citep{pptod}. Nevertheless, there are certain discrepancies between the forms of the two tasks. In particular, TOD can be divided into three sub-tasks: dialogue state tracking, dialogue act decision making, and response generation. The ODD, however, contains only dialogue understanding and response generation modules. Therefore, we start by aligning the data schema and training tasks for TOD and ODD.

For a particular turn of conversation $t$ in a multi-turn dialogue, define the user input as $U_t$ and the system-generated responses of different styles $s\in S$ as $R_t^s$. To model the dialogue history for multiple turns, HiS-Dialog takes all historical turns of conversation as context and generates the belief state $B_t$:
\begin{align}
    C_t&=[U_0, R_0^S,\cdots,U_t], \\%\label{eq:importance}
    B_t&=\textrm{HiS-Dialog}(C_t).
\end{align}

Depending on the dialogue belief state $B_t$, the result $D_t$, which satisfies the $B_t$ restriction, can be retrieved from the database. Then the previously obtained $C_t$, $B_t$ and $D_t$ are aggregated to generate the dialogue action $A_t$:
\begin{align}
    A_t=\textrm{HiS-Dialog}([C_t,B_t,D_t]).
\end{align}

We added act $[chat]$ to the behavioral decision space for the ODD scenario to indicate the system decision for open-domain chit-chat. The information obtained from the individual subtasks can thus be aggregated into a sequence of dialogue messages, termed $\textrm{Dial-INFO}=[C_t,B_t,D_t,A_t]$, and is adapted to generate system responses $R_t^s$ with diverse styles:
\begin{align}
    R_t^S=\textrm{HiS-Dialog}(\textrm{Dial-INFO}).
\end{align}

\subsection{Style Modeling}
\subsubsection{Style Latent Variables}
To alleviate the negative style transfer in hybrid dialogue systems, we should first extract the higher-order style information embedded in the dialogue before the style-controlled text generation. Therefore we utilize the encoder-decoder architecture of the pre-trained model T5 \citep{t5} as the backbone model. Then, based on the pre-trained T5, we further introduced latent variables for modeling the styles of different responses. The dialogue information Dial-INFO is first encoded into sentence representations via HiS-Dialog's encoder:
\begin{align}
    h=\textrm{HiS-Dialog}_\textrm{ENC}(\textrm{Dial-INFO}).
\end{align}
A mapping function is subsequently applied to convert the sentence representation $h$ into the latent variable $z$ containing implicit style information:
\begin{align}
    p_\theta(z|h) = \mathcal{N}(\mu(h),\sigma(h)),
\end{align}

\noindent where $\mathcal{N(\mu,\sigma)}$ is a normal distribution with mean $\mu$ and covariance $\sigma$, attained by a multi-layer feed-forward network with different parameters $\theta$. The model can then generate different styles of dialogue responses $p_\theta(R^s|h,z)$ from the different latent variables $z$ sampled from the distribution $p_\theta(z|h)$.

However, since the estimation of the true posterior distribution $p_\theta(z|h,R^s)$ is intractable, the variational posterior distribution $q_\phi(z|h,R^s)$ is introduced to approximate the true posterior distribution. The procedure can be trained by maximizing the variational lower bound on the marginal likelihood as follows:

\begin{equation}
    \begin{split}
            \log p_\theta(R^s|h)&\ge-\textrm{KL}[q_\phi(z|h,R^s)\|p_\theta(z|h)] \\
            &+\mathbb{E}_{q_\phi(z|h,R^s)}\log p_\theta(R^s|z,h),
    \end{split}
\end{equation}

\noindent where the first term $\textrm{KL}$ denotes optimizing the KL divergence $\mathcal{L}_{\textrm{KL}}$ between the prior and posterior distributions, the latter term represents the maximum likelihood estimate $\mathcal{L}_{\textrm{MLE}}$ of the generated responses, and $\phi$ denotes the variational parameters of the multi-layer feed-forward network for estimating the posterior distribution.

\subsubsection{Style Disentanglement with Contrast}
The style information obtained using only latent variables is not attribute-specific oriented, i.e., all possible styles are entangled. It cannot obtain a specific style variable for a particular task \citep{vgvae}. Therefore, a regularization term is introduced to decouple the styles in the latent space. Specifically, for multiple latent variables representing different text styles sampled in a batch, we expect a certain degree of differentiation between latent variables and retain information on the topic and domain consistency, and therefore introduce a  contrastive loss $\mathcal{L}_{\textrm{CL}}$ with margin. In addition, for the presence or absence of explicit information on text style labels, we design both supervised and self-supervised forms of contrastive loss for different datasets.

For the supervised form, a triplet $(z_a,z_p,z_n)$ can be constructed based on the text style labels \citep{triplet_loss}, where $z_a$ denotes the latent variable acting as the anchor, $z_p$ indicates the positive latent variable with the same text style as the anchor, and $z_n$ represents the negative latent variable with a different text style. For all triples in a batch, the contrastive loss can be calculated as:
\begin{align}
    \mathcal{L}_{\textrm{CL}}&=\sum_{i=1}^{|triplets|}\max(0,\lambda+dis),\label{eq:cl_loss} \\
    dis&=d(z_a,z_p)-d(z_a,z_n),\\
    d(z_i,z_j)&=1-\frac{z_iz_j}{\|z_i\|\|z_j\|},
\end{align}
where $d(\cdot)$ denotes the function used to measure the cosine distance between two latent variables, and $\lambda$ indicates the margin between the positive and negative samples.

Furthermore, a self-supervised approach is designed to construct positive and negative samples for the datasets without style labels. Specifically, the posterior distribution $q_\phi(z|h,R^s)$ is re-parameterised twice to obtain $z$ and $\hat{z}$. Thus, for an in-batch contrastive loss, two batches of latent variables at the same index are mutually positive samples. The other latent variables of the different indexes in the batch are mutually negative samples. Then, the triples for computing the contrastive loss are obtained from the two spliced latent variables, $\tilde{triplet}\leftarrow[z,\hat{z}]$, and the contrastive loss is computed with Eq. \ref{eq:cl_loss} for $\tilde{triplet}$.

\subsection{Generation with Style Prefix}
Once decoupled style latent variables are obtained, it is essential to consider how the latent variables can be more fully integrated with the pre-trained model to steer the generation of varying text styles. First, the dialogue history information Dial-INFO is encoded by the HiS-Dialog encoder to receive the context representation $h$ and is re-parameterized to obtain the style variable $z$. Next, the key and value of the decoder in computing the cross-attention are the sum of the context representation $h$ and the latent variable $z$.
For cross-attention in the decoder, the process is computed as:

\begin{align}
    \textrm{K}=\textrm{V}=z\oplus h.
\end{align}
    
The distribution of $\textrm{K},\textrm{V}$ differs from the pre-training stage due to the introduction of $z$ and the relatively small portion of stylized data, which leads the original $\textrm{W}^K,\textrm{W}^V$ to focus primarily on the text content and neglect the stylized information. Therefore, to further enhance the decoder's procedure for different text generation styles using style latent variables, we concatenated a small number of trainable vectors $\textrm{P}^K$ and $\textrm{P}^V$ in front of the keys and values of the attention head as a continuous instruction:

\begin{gather*}
    head_i=\textrm{AT}(\textrm{QW}_i^Q,[\textrm{P}_i^K,\textrm{KW}_i^K],[\textrm{P}_i^V,\textrm{VW}_i^V]), \nonumber\\
    \textrm{MHA(Q,K,V)}=\textrm{Concat}(head_1,\cdots,head_n)\textrm{W}^O, \nonumber    
\end{gather*}

\noindent where AT denotes attention computation and MHA indicates multi-head attention \citep{survey_trans}. Thus, for various styles of responses $r_j\in R^s$ of length $N$, the decoding process for each time step $j\in \{1,\cdots,N\}$ is:

\begin{align}
    h_j^d=\textrm{HiS-Dialog}_{\textrm{DEC}}(h,z,h_{<j}^d),
\end{align}

\noindent where $h_j^d$ denotes the decoder hidden state of each token output during decoding. Then the entire sentence is generated based on the encoder hidden state and the representation generated in the previous decoding time step:

\begin{align}
    p(r_j|r_{j-1},z)=\textrm{Softmax}(\textrm{W}_d\cdot h_j^d),
\end{align}

\noindent where $\textrm{W}_b$ is the parameter matrix that maps the decoder output to the vocabulary distribution. Finally, the overall training loss is:

\begin{align}
    \mathcal{L}=\mathcal{L}_{\textrm{MLE}}+\alpha\mathcal{L}_{\textrm{KL}}+\beta\mathcal{L}_{\textrm{CL}},
\end{align}

\noindent where $\alpha$ and $\beta$ are coefficients that control the scale of different losses.

\begin{table*}[t]
\centering
\resizebox{0.9\textwidth}{!}{
\setlength{\tabcolsep}{1.2mm}
\begin{tabular}{lcccccccc}
\toprule
\textbf{Dataset}     & \multicolumn{8}{c}{\textbf{FusedChat}}                              \\
\midrule
           & \multicolumn{4}{c}{\textbf{TOD}}     & \multicolumn{4}{c}{\textbf{ODD}}        \\
\cmidrule(lr){2-5}\cmidrule(lr){6-9}
\textbf{Metrics}     & Inform & Success & BLEU  & Combined Score & DIST-1 & DIST2 & DIST-3 & BLEU \\
\midrule
MinTL \citep{mintl}     & 80.80  & 74.40   & 16.20 & 93.80   & 0.03   & 0.15  & 0.28   & 9.02 \\
PPTOD \citep{pptod}     & 90.40  & 82.50   & 14.32 & 100.77  & 0.03   & 0.14  & 0.26   & 9.00 \\
MTTOD \citep{mttod}     & 90.50  & 82.50   & 17.23 & 103.73  & 0.04   & 0.16  & \textbf{0.36}   & 9.07 \\
T5 \citep{t5}       & 90.30  & 81.90   & 17.48 & 103.58  & 0.03   & 0.15  & 0.29   & 9.26 \\
T5-CVAE \citep{t5_cvae}   & 91.60  & 82.90   & 16.88 & 104.13  & 0.04   & 0.15  & 0.28   & 9.11 \\
\midrule
Hy-Dialog &
  \begin{tabular}[c]{@{}c@{}}\textbf{92.00}\\ ($\pm$0.78)\end{tabular} &
  \begin{tabular}[c]{@{}c@{}}\textbf{84.40}\\ ($\pm$0.65)\end{tabular} &
  \begin{tabular}[c]{@{}c@{}}\textbf{17.58}\\ ($\pm$0.82)\end{tabular} &
  \begin{tabular}[c]{@{}c@{}}\textbf{105.73}\\ ($\pm$0.93)\end{tabular} &
  \begin{tabular}[c]{@{}c@{}}\textbf{0.04}\\ ($\pm$0.01)\end{tabular} &
  \begin{tabular}[c]{@{}c@{}}\textbf{0.17}\\ ($\pm$0.04)\end{tabular} &
  \begin{tabular}[c]{@{}c@{}}0.31\\ ($\pm$0.08)\end{tabular} &
  \begin{tabular}[c]{@{}c@{}}\textbf{9.51}\\ ($\pm$0.26)\end{tabular} \\
\bottomrule
\end{tabular}
}
\caption{Overall results of HiS-Dialog compared to other methods under two subtasks of FusedChat.}
\label{tab:main_results}
\end{table*}

\section{Experiments}
\subsection{Evaluation Datasets}
We conducted experiments in end-to-end settings under three dialogue datasets containing one hybrid dialogue dataset, FusedChat, and two task-oriented datasets, MultiWOZ2.0, and MultiWOZ2.1.

\noindent \textbf{FusedChat} \citep{fusedchat} This dataset expands or rewrites each conversation based on the task-oriented task. By annotating the original TOD data with additional rounds of context-semantically relevant open-domain dialogue, a conversation can contain multiple dialogue patterns while increasing the average number of turns by 5.8.

\noindent \textbf{MultiWOZ} \citep{multiwoz, multiwoz21} This dataset is one of the most prevalent datasets in task-oriented dialogue systems, collected via Wizard-of-Oz, and contains a total of 8438/1000/1000 multiturn dialogues. Among them, MultiWOZ 2.0 is the initially proposed multi-domain goal-directed task-oriented dataset, while MultiWOZ 2.1 is a version with several buggy annotations fixed. We evaluate both datasets to assess the robustness of the model. The reason for analyzing these datasets is that they do not provide text style information but implicitly reflect the text style of different domains or themes, allowing the method's effectiveness to be validated in a self-supervised setting.

\subsection{Metrics}
For the assessment metrics in task completion, we utilized the evaluation scripts provided by the dataset to evaluateitep{multiwoz} of the experimental results. In particular, \textbf{Inform} was used to evaluate whether the system generated the entities mentioned by the user, \textbf{Success} measured whether all the slots requested by the user were fulfilled, and \textbf{BLEU} assessed the fluency of the system in generating responses. The final combined score is calculated as $\textrm{Combined = (Inform + Success) * 0.5 + BLEU}$.

To assess the efficacy of open-domain response generation, we calculated the number of unique 1/2/3-grams in a sentence using Distinct-1/2/3 \citep{distinct} to measure the diversity of responses generated by the system.

Moreover, we calculated BLEU for both task-oriented and open-domain dialogue responses to evaluate the model's linguistic quality in generating task-related and open-domain responses.

\subsection{Implementation Setup}
HiS-Dialog and other baselines based on pre-trained models are implemented with HuggingFace's Transformers. We employ AdamW as the optimizer and configure the warmup rate to 0.1. For FusedChat, the learning rate is 6e-4 and the batch size is set to 12 for 12 epochs. For MultiWOZ, we set the learning rate 5e-4, epoch 10, and batch size 12. All experiments were performed on a GeForce RTX 3090 GPU (24G), and the mean of the results from three different random seeds was selected as the final result.

\subsection{Compared Baselines}
We conducted comparative experiments for the three benchmark datasets with the following robust and relatively new end-to-end dialogue systems based on pre-trained models.

\emph{Encoder-decoder architectures}: (1) \textbf{DAMD} \cite{damd} augments dialogue actions with additional data to provide more diverse responses. (2) \textbf{MinTL} \citep{mintl} generates dialogue states and responses in succession with two decoders. (3) \textbf{PPTOD} \citep{pptod} models multiple dialogue tasks uniformly as generative tasks by constructing task-specific prompts. (3) \textbf{MTTOD} \citep{mttod} introduces an additional span prediction task on the encoder side on top of the two-decoder structure. (5) \textbf{T5-CVAE} \citep{t5_cvae} models additional latent variables on top of T5. (6) \textbf{DoTS} \citep{dots} reduces the requirement for historical dialogue length by additionally modelling the domain state.

\emph{Auto-aggressive architectures}: (1) \textbf{SimpleTOD} \citep{simpletod} models all tasks in TOD as autoregressive generation with GPT2 \citep{gpt2}. (2) \textbf{SOLOIST} \citep{soloist} further pre-trains the model with heterogeneous dialogue datasets. (3) \textbf{UBAR} \citep{ubar} additionally adding dialogue states, database information, and dialogue actions to the dialogue history.

\begin{table*}[t]
\centering
\resizebox{0.95\textwidth}{!}{
\setlength{\tabcolsep}{1.2mm}
\begin{tabular}{lcccccccc}
\toprule
\textbf{Dataset}    & \multicolumn{4}{c}{\textbf{MultiWOZ 2.0}}   & \multicolumn{4}{c}{\textbf{MultiWOZ 2.1}}   \\
\cmidrule(lr){1-1}\cmidrule(lr){2-5}\cmidrule(lr){6-9}
\textbf{Metrics}    & Inform & Success & BLEU  & Combined Score & Inform & Success & BLEU  & Combined Score \\
\midrule
DAMD \citep{damd}  & 76.33  & 60.40   & 16.60 & 84.97   & -  & -   & - & -   \\
SimpleTOD \citep{simpletod}  & 84.40  & 70.10   & 15.01 & 92.26   & 85.00  & 70.50   & 15.23 & 91.98   \\
DoTS \citep{dots}       & 86.59  & 74.14   & 15.06 & 95.43   & 86.65  & 74.18   & 15.90 & 96.32   \\
SOLOIST \citep{soloist}    & 85.50  & 72.90   & 16.54 & 95.74   & -      & -       & -     & -       \\
MinTL-BART \citep{mintl} & 84.88  & 74.91   & 17.89 & 97.79   & -      & -       & -     & -       \\
UBAR  \citep{ubar}   & 95.40  & 80.70   & 17.00 & 105.10  & \textbf{95.70}  & 81.80   & 16.50 & 105.25  \\
MTTOD \citep{mttod}      & 91.00  & 82.60   & \textbf{21.60} & 108.30  & 91.00  & 82.10   & \textbf{21.00} & 107.50  \\
PPTOD \citep{pptod}      & 89.20  & 79.40   & 18.62 & 102.92  & 87.09  & 79.08   & 19.17 & 102.26  \\
T5 \citep{t5}         & 90.70  & 81.30   & 18.94 & 104.94  & 91.10  & 82.00   & 18.34 & 104.89  \\
\midrule
HiS-Dialog  & \textbf{92.85}  & \textbf{84.30}   & 20.12 & \textbf{108.70}  & 92.30  & \textbf{83.90}   & 19.76 & \textbf{107.86}  \\
\bottomrule
\end{tabular}
}
\caption{HiS-Dialog results in an end-to-end paradigm on two different versions of the MultiWOZ dataset.}
\label{tab:multiwoz}
\end{table*}

\subsection{Overall Performance}
Table \ref{tab:main_results} illustrates the results of HiS-Dialog with other baselines on the hybrid dialogue dataset FusedChat. It is evident from the results that our approach achieves state-of-the-art results in both the TOD segment and the ODD segment. In the TOD part, HiS-Dialog's Inform and Success scores for task completion improved by 0.4 and 1.5 points, respectively, compared to the  T5-CVAE \citep{t5_cvae}. In the ODD section, HiS-Dialog obtained the latest results on both Distinct-2 and BLEU, indicating our method achieves higher diversity and quality of open domain text. Combining the results of TOD and ODD, it is observed that HiS-Dialog achieves a significant performance improvement in both subtasks compared to the other baselines. The results also indirectly indicate that HiS-Dialog is more capable of mitigating the negative transfer phenomenon between tasks, thus improving the overall effectiveness of the model.

We evaluated the performance of the proposed method in a self-supervised setting without text style labels, and exploratory experiments were done using MultiWOZ with the different implicit domain or topic styles. In addition, we assume that \texttt{inform} and \texttt{success} are utilized to signify task completion, and \texttt{BLEU} is employed to measure the quality of text generation. Hence, a rise in each of the two metrics indicates that the negative style transfer issue has been alleviated. Table \ref{tab:multiwoz} indicates that HiS-Dialog achieved the latest combined scores on the MultiWOZ 2.0 and 2.1 datasets. Of these, MTTOD has the relatively highest BLEU scores on both datasets since MTTOD employs two decoders for dialogue states and response generation, respectively, but with a 50\% increase in the overall number of model parameters. Nonetheless, HiS-Dialog achieves the latest combined scores with only contrastive latent variable and style prefix that adds a small number of parameters, demonstrating that our approach mitigates the effects of negative style transfer, resulting in enhancements to both task completion rate and text generation quality.

\begin{table}[t]
\centering
\resizebox{1\columnwidth}{!}{
\begin{tabular}{lllll}
\toprule
          & Inform       & Success      & BLEU         & Comb          \\
\midrule
HiS-Dialog & 90.20        & 81.80        & 18.60        & 104.60        \\
+CLV       & 91.20 ($\uparrow$1.00) & 83.20 ($\uparrow$1.40) & 19.05 ($\uparrow$0.45) & 106.25 ($\uparrow$1.65) \\
+SP       & 92.50 ($\uparrow$2.30) & 83.75 ($\uparrow$1.95) & 19.63 ($\uparrow$1.03) & 107.76 ($\uparrow$3.16) \\
+ CLV+SP   & 92.60 ($\uparrow$\textbf{2.40}) & 84.60 ($\uparrow$\textbf{2.80}) & 19.79 ($\uparrow$\textbf{1.19}) & 108.39 ($\uparrow$\textbf{3.79}) \\
\bottomrule
\end{tabular}
}
\caption{Efficacy of contrastive latent variables (CLV) and style prefix (SP) in mitigating negative style transfer.}
\label{tab:ablation}
\end{table}

\begin{table}[t]
\centering
\resizebox{0.98\columnwidth}{!}{
\begin{tabular}{lcccc}
\toprule
          & \multicolumn{2}{c}{\textbf{FusedChat}} & \multicolumn{2}{c}{\textbf{MultiWOZ}} \\
\cmidrule(lr){2-3}\cmidrule(lr){4-5}
Metric    & Informative $\downarrow$         & Differential $\downarrow$       & Informative $\downarrow$        & Differential $\downarrow$       \\
\midrule
PPTOD     & 3.27           & 3.57         & 2.75          & 3.26         \\
MTTOD     & 2.89           & 2.56         & 2.71          & \textbf{2.24}         \\
HiS-Dialog & \textbf{2.16}           & \textbf{2.54}         & \textbf{2.63}          & 2.39         \\
\cmidrule(lr){1-1}\cmidrule(lr){2-3}\cmidrule(lr){4-5}
Golden    & 1.68           & 1.23         & 2.01          & 2.11         \\
\bottomrule
\end{tabular}
}
\caption{The responses generated by the different approaches were ranked by human evaluation. Informative indicates the informative adequacy of the generated sentences, Differential denotes the distinguishability of the different text styles, and Golden refers to the correct responses used for reference.}
\label{tab:human_eval}
\end{table}

\begin{figure}[t]
\centering
\includegraphics[width=0.98\columnwidth]{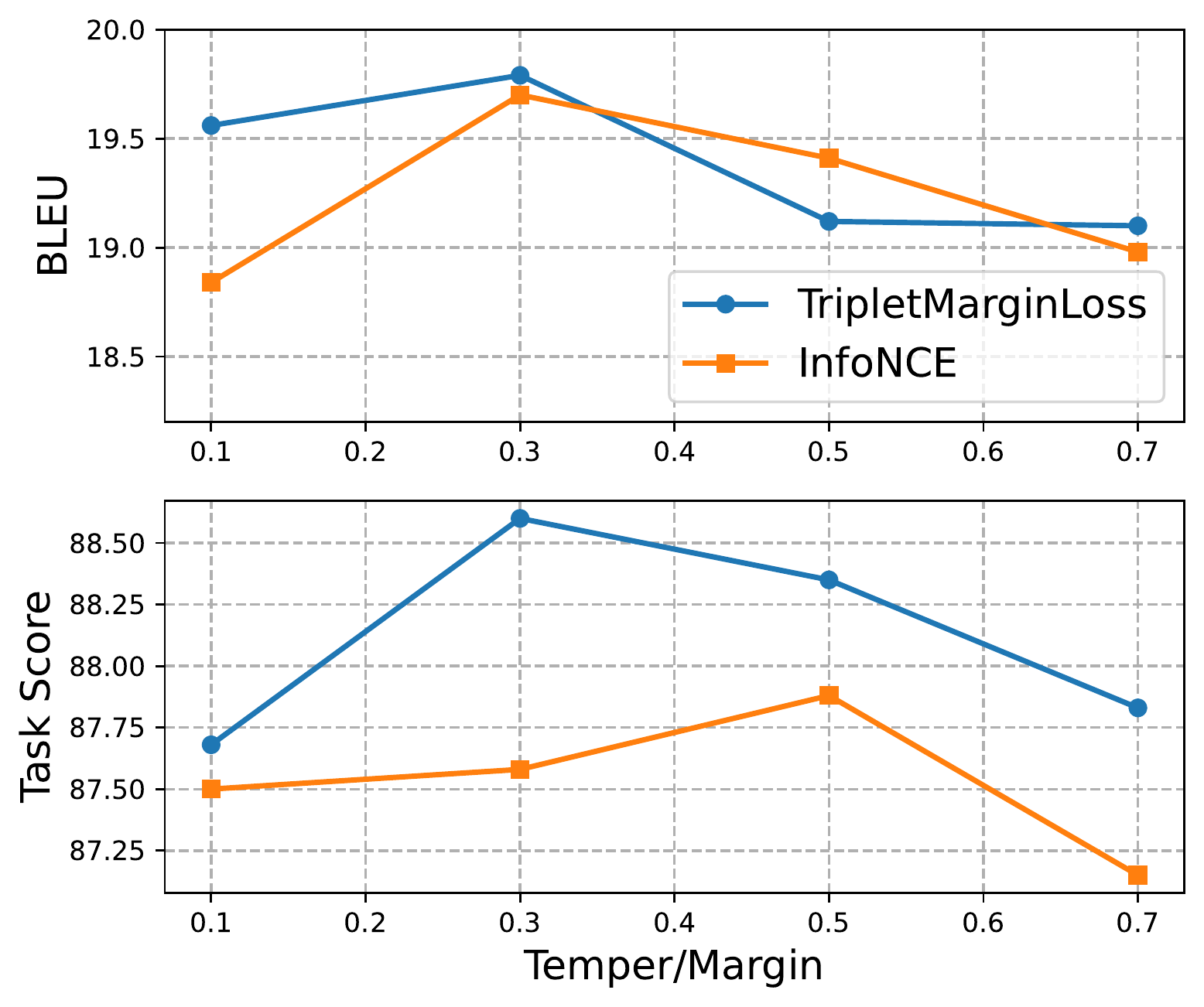} % Reduce the figure size so that it is slightly narrower than the column.
\caption{The effect of various contrastive loss functions and varying margin/temperature coefficients on performance.}
\label{fig: different_cl}
\end{figure}

\begin{table}[t]
\centering
\resizebox{0.95\columnwidth}{!}{
\setlength{\tabcolsep}{1mm}
\begin{tabular}{@{}lccccc@{}}
\toprule
\quad Prefix type                    & \multicolumn{1}{l}{Prefix-length} & Inform         & Success        & BLEU           & Combine         \\ 
\midrule
\multirow{4}{*}{\quad Enc+Dec+Cross} & 30                                & 92.45          & 83.30          & 19.35          & 107.22          \\
                               & 50                                & 92.60          & \textbf{84.60} & \textbf{19.79} & \textbf{108.39} \\
                               & 70                                & 91.50          & 83.20          & 19.52          & 106.87          \\
                               & 100                               & 92.60          & 84.60          & 19.32          & 107.92          \\ 
\midrule
\quad Dec                            & 50                                & 91.60          & 83.40          & 19.61          & 107.11          \\ 
\midrule
\quad Enc+Dec                        & 50                                & \textbf{93.00} & 84.35          & 19.60          & 108.27          \\ 
\bottomrule
\end{tabular}
}
\caption{Impact of different lengths and positions of style prefix on the effectiveness of hybrid dialogue systems.}
\label{tab:prefix_len}
\end{table}

\subsection{Effect on Mitigating Negative Style Transfer}
\subsubsection{Ablation}
This section experiments on the effects of latent variable contrastive learning and style prefix in HiS-Dialog on mitigating the negative style transfer problem in hybrid dialogue systems. The experimental results in Table \ref{tab:ablation} illustrate that adding additional contrastive loss to the optimization process in the latent space for constraint leads to a 1.0 and 1.4 point improvement in task success rate and text generation quality, thus corroborating the gain of contrastive learning in mitigating negative style transfer.

Furthermore, by additionally introducing a style prefix into the attention calculation, the impact of negative style transfer of responses of different styles can also be mitigated, thus significantly improving the overall performance of the hybrid dialogue system. Lastly, adding the two mechanisms together improves the overall score of the hybrid dialogue system by 3.79 points, suggesting that style prefixing enables further style-controlled text generation by fully using the well-defined style variables obtained from latent space modeling.

\subsubsection{Human Evaluation}
To further assess the quality of the generated responses from different models, we evaluated the informativeness and distinguishability of their generated responses on FusedChat and MultiWOZ. We randomly selected 50 dialogues from both datasets and invited five domain-related practitioners to rank the responses generated by these methods. Table \ref{tab:human_eval} clearly illustrates that the informativeness of HiS-Dialog-generated responses receives the highest-ranking scores among the baselines in both datasets. Regarding distinguishability of response styles, HiS-Dialog obtains the highest ranking in the hybrid dialogue dataset FusedChat, while MTTOD obtains the highest-ranking score on MultiWOZ. Also, FusedChat provides explicit text style labels, allowing HiS-Dialog's contrastive learning to learn better style variables. In contrast, MultiWOZ does not provide style labels, so the two decoder architectures of MTTOD can generate more distinguishable sentences.

\subsection{Analysis}
\subsubsection{Efficacy of Latent Variable Contrastive Learning}
We explored the efficacy of latent variable contrastive learning with two forms of loss functions and varying sizes of hyperparameters. TripletMarginLoss \citep{triplet_loss} is the triplet loss employed in our method, and InfoNCE \citep{simcse} is another widely used loss function for contrastive learning. The hyperparameter explored in the triplet loss is the relative distance margin between positive and negative samples, and the temperature coefficient $\tau $ is analyzed in InfoNCE.

To explore the influence of different settings on task completion and generation quality, we define \texttt{Task Score} = (\texttt{Inform} + \texttt{Success}) * 0.5 for task completion and BLEU to assess the quality of text generation. As illustrated in Figure \ref{fig: different_cl}, the overall results of triplet loss compared to InfoNCE are generally better under different hyperparameters. A possible explanation is that InfoNCE only pulls in the distance between the anchor point and the positive sample. In contrast, triplet loss further considers the relative distance between the anchor point and the negative samples.
Also, 0.3 was also chosen as the best margin value in our method.

\subsubsection{Analysis of Style Prefix}
We investigate the effect of different lengths and positions of style prefixes on the quality of text generation for style controllable. From Table \ref{tab:prefix_len}, we can observe that HiS-Dialog achieves the highest combined score when the size of the style prefix is set at 30. As the prefix length increases, the system performance degrades to some extent, suggesting that excessively long style prefixes can cause the model to over-fit to style-specific information, thereby reducing the diversity of the generation.
 Style prefix being added to both the encoder side and the cross-attention computation results in better performance than adding style prefix to the decoder side alone. It indicates that the style prefix on the encoder side can further extract style-related information, and style prefixes in the cross-attention computation can better guide the decoder's controlled style generation with the extracted style information.

\section{Conclusion}
We propose a novel latent variable encoder-decoder model, HiS-Dialog, which incorporates contrastive loss to constrain the latent space for better control of varying styles of text generation and to mitigate negative style transfer. We also introduce a style prefix to fully exploit the modeled style latent variables to guide the process of generating diverse styles of responses. Empirical results on three dialogue datasets demonstrate that our approach mitigates the occurrence of negative transfer in a better way than previous baselines and achieves improvements in several aspects.

\section{Acknowledgments}
% We appreciate the reviewers' diligence, rigor, responsibility, and insightful remarks that will improve our work.
This work was supported by the National Key Research and Development Program of China (No.2020AAA0108700) and National Natural Science Foundation of China (No.62022027).

\bibliography{main}

\end{document}